\documentclass{mva_style}
\usepackage{graphicx}
\usepackage{color}
\usepackage{cite}
\usepackage{amsmath,amssymb,amsfonts}
\usepackage{algorithmic}
\usepackage{textcomp}
\usepackage{xcolor}
\usepackage{multirow}

\usepackage{enumitem}
\usepackage{physics}
\usepackage{url}

\finalcopy

\usepackage{verbatim}

\newcommand{\tabincell}[2]{\begin{tabular}{@{}#1@{}}#2\end{tabular}}
\begin{document}
\title{Data-driven Head Motion Generation through Natural Gaze-Head Coordination%
\thanks{\copyright~2025 IEEE. Personal use of this material is permitted. Permission from IEEE must be obtained for all other uses, in any current or future media, including reprinting/republishing this material for advertising or promotional purposes, creating new collective works, for resale or redistribution to servers or lists, or reuse of any copyrighted component of this work in other works. Accepted version of: X.~Liu, Y.~Wen, and Y.~Sugano, ``Data-driven Head Motion Generation through Natural Gaze-Head Coordination,'' in \textit{Proc. 19th Int. Conf. on Machine Vision and Applications (MVA)}, Kyoto, Japan, Jul.\ 2025, pp.~1--5. DOI: \texttt{10.23919/MVA65244.2025.11175115}. Final published version: \protect\url{https://doi.org/10.23919/MVA65244.2025.11175115}.}}

\author{
  Xiaohan Liu, Yilin Wen, and Yusuke Sugano\\
  Institute of Industrial Science, The University of Tokyo\\
  Komaba 4-6-1, Tokyo, Japan\\
  {\tt \{xhliu, fylwen, sugano\}@iis.u-tokyo.ac.jp}\\
}

\maketitle

\section*{\centering Abstract}
\textit{
  We present the first data-driven approach to model temporal gaze-head coordination from large-scale in-the-wild facial videos.
  To obtain training data for generalizable learning, we propose an automatic pipeline that extracts natural yet diverse gaze and head motions with off-the-shelf appearance-based gaze estimators.
  To capture the probabilistic correlation and temporal dynamics of gaze-head coordination, we build our model on a generative conditional Variational Autoencoder for plausible yet diverse gaze-conditioned head motion generations.
  We further apply our framework to gaze-controlled facial video generation, where we enable video generation with natural and realistic head motion correlated to the input gaze - an aspect that has not been emphasized before.
  Human evaluation and quantitative comparisons demonstrate our method's effectiveness and validate our design choices, with evaluators showing statistically significant preference for our approach over baseline methods.
}

\section{Introduction}

Human visual attention relies on a sophisticated coordination between eye gaze and head movements, where head orientation naturally complements gaze direction in a synchronized manner~\cite{pelz2001coordination}. Modeling this gaze-head coordination is fundamentally important for understanding human oculomotor mechanisms and enables critical applications in multimedia, including virtual avatar animation, talking head synthesis, and interactive digital humans. These applications require naturalistic head movements that appropriately complement gaze direction to avoid rigid or unnatural appearances. For example, recent advances in gaze redirection~\cite{zheng2020self,jin2023redirtrans,ruzzi2023gazenerf} have progressed in synthesizing gaze-controlled facial images, these approaches typically treat head pose as an independent control variable, failing to capture the intrinsic coordination between gaze and head movements. This limitation results in stiff and unnatural head motions, particularly evident during dynamic gaze transitions.

To model gaze-head coordination, previous research has predominantly relied on data collected from controlled laboratory settings~\cite{vercher1991eye,ackerley2011interaction}, mobile eye trackers~\cite{kothari2020gaze}, or VR headsets~\cite{hu2019sgaze,hu2020dgaze}. However, these approaches face inherent limitations due to specialized hardware requirements and restricted data collection environments, which impede effective modeling of the complex probabilistic relationships and temporal dynamics that characterize natural gaze-head coordination patterns.

In this paper, we address the challenge of modeling gaze-head coordination through a two-pronged approach. First, we develop an automatic pipeline that efficiently extracts diverse and natural gaze and head motion data from large-scale in-the-wild videos. Second, we introduce the first data-driven framework that accurately captures both the probabilistic nature of the distributional mapping and the temporal dynamics of gaze-head coordination, enabling the generation of plausible and diverse head motions that naturally follow input gaze sequences.

For data acquisition, we leverage recent advances in appearance-based gaze estimation by employing an off-the-shelf gaze estimator~\cite{Qin_2022_CVPR} on large-scale in-the-wild facial videos. This approach circumvents the constraints associated with specialized hardware setups while obtaining accurate, natural, and diverse gaze and head pose data across various subjects and environments. Using this collected data, we construct a generative conditional Variational Autoencoder (cVAE) that effectively captures the probabilistic nature and temporal dynamics of gaze-head coordination. Our model generates diverse head motions conditioned on input gaze sequences, with the VAE latent space encoding the distributional mapping of gaze-head coordination patterns. We further incorporate a Gated Recurrent Unit (GRU) to model temporal dynamics and utilize historical generation as context to enable smooth long-term generation in an autoregressive manner.

We evaluate our framework on gaze and head motion data extracted from the CelebV-Text dataset~\cite{yu2022celebvtext}, with additional application to facial video generation by integrating our generated head motions with a pre-trained face image generator. This represents the first focused attempt to explicitly model gaze-head coordination for facial synthesis applications. Human evaluation with expert assessors demonstrates statistically significant preference for our method over baseline approaches in terms of naturalness and coordination with gaze. Diverse quantitative and qualitative evaluations further confirm the plausibility and diversity of our generated head motions across both direct head motion analysis and facial video generation tasks.

Our contributions can be summarized as follows: (1) The first data-driven approach for modeling gaze-head coordination; (2) An automatic pipeline for efficiently extracting natural and diverse gaze-head coordination data from in-the-wild facial videos; (3) A cVAE-based framework that effectively captures both the probabilistic nature and temporal dynamics of gaze-head coordination; and (4) Extensive evaluation demonstrating our method's effectiveness in generating plausible yet diverse gaze-conditioned head motions.

\section{Method}

\begin{figure}[t]
    \centering
    \includegraphics[width=0.9\columnwidth]{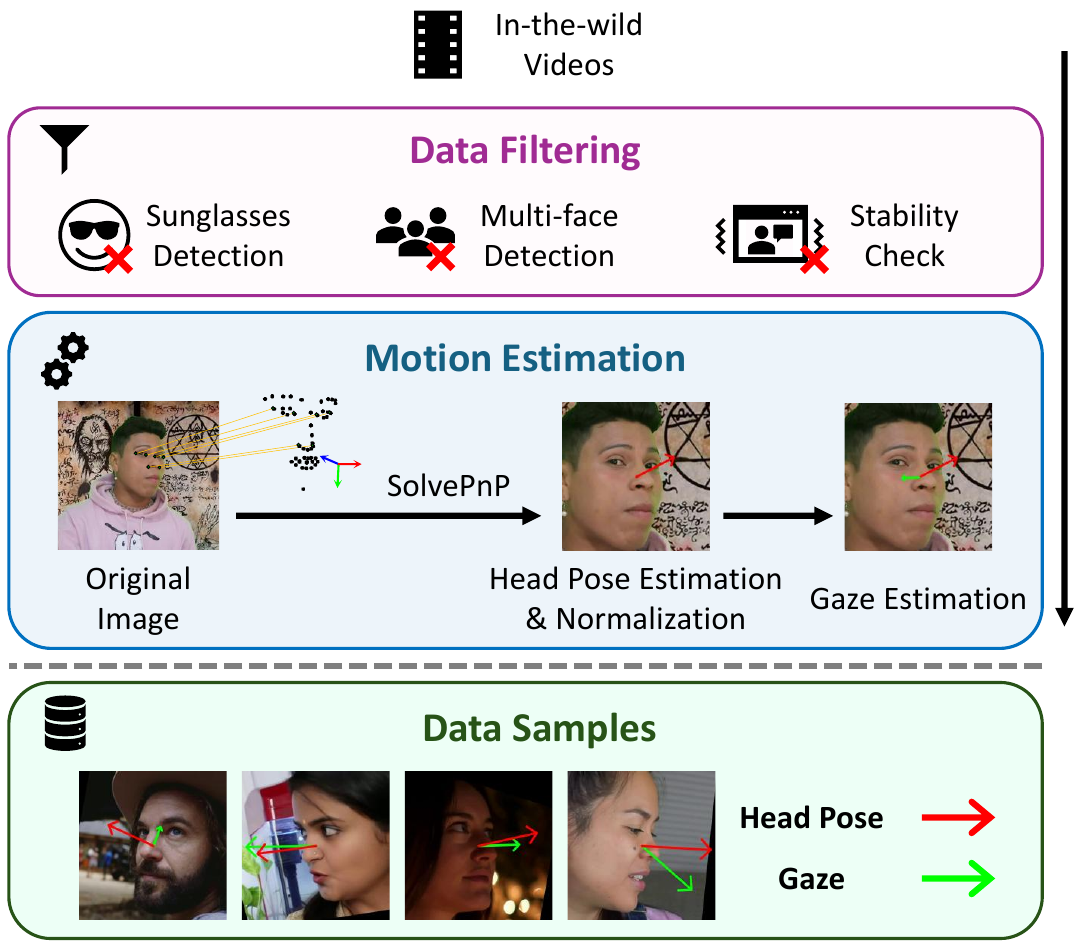}
    \caption{Our automatic pipeline to extract gaze-head coordination data from in-the-wild videos.}
    \label{fig:overview_data}
    \vspace{-7mm}
\end{figure}

\begin{figure*}[t]
    \centering
    \includegraphics[width=0.8\linewidth]{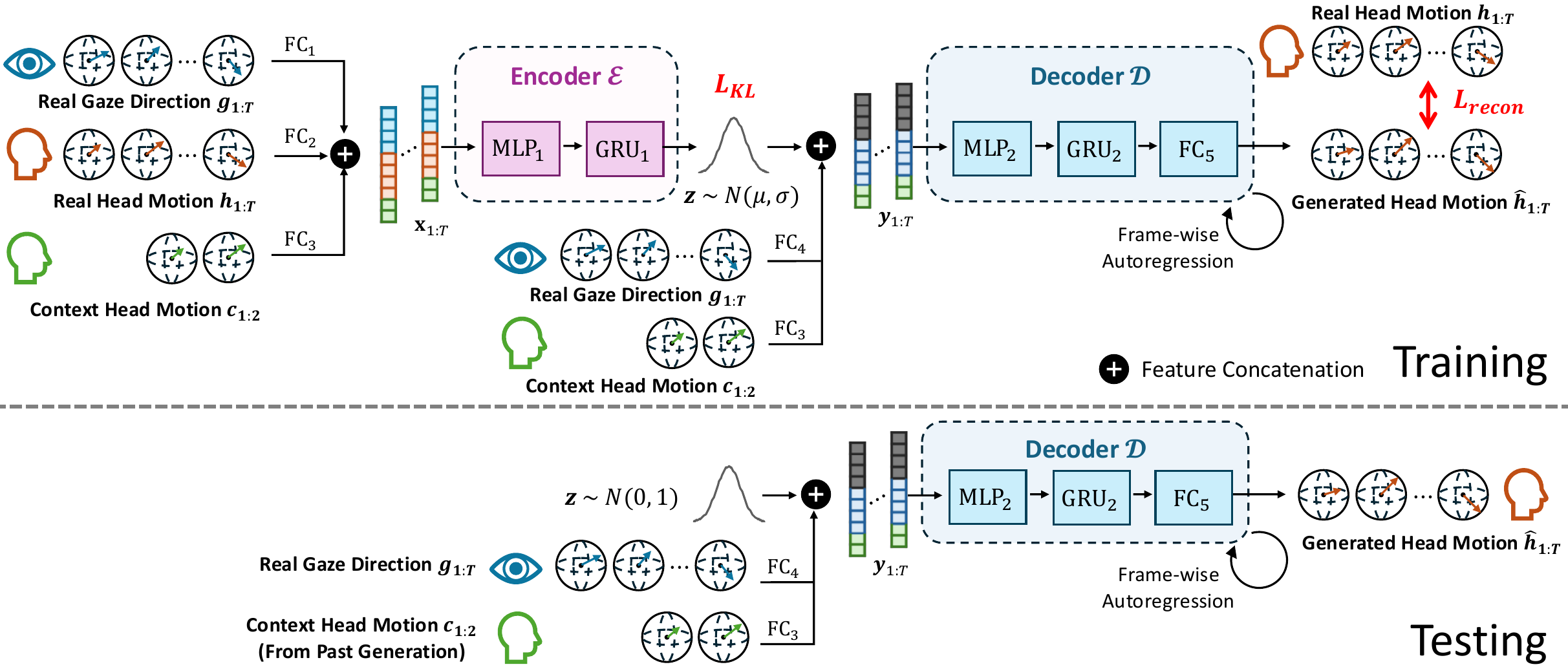}
    \caption{Our cVAE framework for modeling gaze-head coordination during training and testing phases. The model generates head motion $\hat{\vb{h}}$ conditioned on gaze sequences $\vb{g}$ and leverages context $\vb{c}$ from historical generation for smooth long-term results.}
    \label{fig:overview_model}
    \vspace{-7mm}
\end{figure*}

We propose a data-driven solution for gaze-head coordination modeling that consists of three key components: (1) an automatic pipeline to obtain training data from in-the-wild facial videos (Fig.~\ref{fig:overview_data}), (2) a cVAE framework to capture the probabilistic nature and temporal dynamics of gaze-head coordination (Fig.~\ref{fig:overview_model}), and (3) integration with facial image generators for video synthesis.

\subsection{Obtaining Training Data}\label{method:data_collection}
We employ an automatic pipeline to extract gaze and head motions from in-the-wild facial videos, enabling efficient collection of diverse yet natural gaze-head movements without specialized hardware constraints.

As shown in Fig.~\ref{fig:overview_data}, we extract head poses and gaze directions from facial videos using an off-the-shelf appearance-based gaze estimator~\cite{Qin_2022_CVPR,zheng2020self}, which generalizes well across different head poses, subjects, lighting and backgrounds. The estimator first determines the head pose through facial landmark detection~\cite{bulat2017far} followed by PnP-based pose estimation with 3D facial model matching. Following Zhang et al.~\cite{zhang2018revisiting}, the estimator then applies a normalization process that transforms the detected face to a standardized canonical view (with fixed camera parameters), which helps cancel out geometric variations across different head poses and camera positions. Finally, it outputs the gaze direction. We denote gaze direction and head pose as $\vb*{g}_t = [\theta^g_t, \phi^g_t]$ and $\vb*{h}_t = [\theta^h_t, \phi^h_t]$ (pitch and yaw angles).

To ensure data quality, we filter out videos with: (1) sunglasses detected by the Glasses Detector~\cite{Birskus_Glasses_Detector_2024}, (2) multiple faces or frequent face detection failures identified using facial landmarks with Google MediaPipe~\cite{48292}, and (3) unstable camera motion or rapid scene changes identified with PySceneDetect~\cite{Castellano_PySceneDetect}.

\subsection{Modeling Gaze-Head Coordination}\label{method:cvae}

As shown in Fig.~\ref{fig:overview_model}, we build our framework on a conditional Variational Autoencoder (cVAE) to capture both probabilistic correlations and temporal dynamics of gaze-head coordination. Given a sequence of gaze directions $\vb*{g}_{1:T}$, the cVAE generates corresponding head poses $\hat{\vb*{h}}_{1:T}$ that follow natural coordination patterns. 

Both the encoder $\mathcal{E}$ and decoder $\mathcal{D}$ employ Gated Recurrent Units (GRU) to leverage temporal continuity. For long-term generation beyond the sequence length $T$, we implement an autoregressive approach where the head poses generated in the previous iteration ($\hat{\vb*{h}}_{T-1}$ and $\hat{\vb*{h}}_T$) serve as context conditions $\vb*{c}_1, \vb*{c}_2$ for the next iteration, maintaining temporal coherence across segments.

The encoder $\mathcal{E}$ takes concatenated features $\vb*{x}_{1:T}$ from the gaze directions, real head motions, and context history, preprocessed by fully connected layers. It outputs parameters $\mu$ and $\sigma$ for sampling the latent variable $\vb*{z} \sim \mathcal{N}(\mu, \sigma)$, capturing the probabilistic nature of gaze-head coordination.

The decoder $\mathcal{D}$ takes the latent variable $\vb*{z}$, preprocessed gaze inputs, and context history to generate head motions $\hat{\vb*{h}}_{1:T}$ through frame-by-frame autoregression. The GRU's hidden state is iteratively updated to maintain temporal consistency.

Our loss function combines reconstruction loss and KL-divergence:
\begin{equation}
   L = ||\hat{\vb*{h}}_{1:T}-\vb*{h}_{1:T}||_2 + \lambda L_{KL}
\end{equation}
where $L_{KL}$ regularizes the latent space distribution and $\lambda$ balances the two terms.

During inference, we use only the decoder $\mathcal{D}$, sampling the latent variable from $\mathcal{N}(0,1)$. For initial frames, context vectors are set to zero. Long-term generation proceeds autoregressively, with each iteration using head poses from previous generations as context.

\subsection{Implementation Details}\label{method:implementation}

Our model processes sequences at 5 FPS, with a window of 12 frames capturing approximately 2.4 seconds of motion. The cVAE latent has a dimension of $d=128$, with both the encoder and decoder utilizing GRUs to process temporal information. We train the model for 60,000 steps with a batch size of 64 using Adam optimizer~\cite{kingma2014adam} with learning rate $5\times 10^{-5}$. Regularization techniques including KL-divergence weight annealing, context dropout, and feature dropout are employed for stable training.

For facial video generation, we integrate our head motion generator with ST-ED [2] pre-trained on ETH-XGaze [18], enabling synthesis of facial videos with natural gaze-head coordination.

\section{Experiments}

\subsection{Dataset and Evaluation Protocol}
We train and evaluate our approach on the CelebV-Text dataset~\cite{yu2022celebvtext}, which contains diverse in-the-wild facial videos with natural head and gaze behaviors. Following the extraction and filtering pipeline described in Section~\ref{method:data_collection}, we obtain 45,806 training videos and 676 testing videos from unseen subjects. All videos have their frame rate aligned to 25 FPS for consistency.

For evaluation, we use each gaze sequence from the test videos as input and randomly generate 30 head motion sequences per input to assess both plausibility and diversity of our results. Additional video demonstrations are available in the supplementary materials.

\subsection{Evaluation Metrics}
We evaluate using metrics for both plausibility and diversity. For plausibility, we measure \textbf{Angular Error}, which captures the 3D angular difference between generated and real head orientations. We also compute \textbf{Correlation Coefficient} to assess temporal correlation for pitch and yaw dimensions separately. \textbf{Average Variance Error (AVE)} quantifies the difference between variances of generated and real head poses, while \textbf{Smoothness} uses the third derivative of head pose sequences to measure motion continuity. For diversity assessment, we employ \textbf{Average Pairwise Distance (APD)}~\cite{yuan2020dlow}, which measures variation among head motions generated from the same input.

To validate perceptual quality, we conducted a human evaluation study with four expert evaluators who assessed pairs of generated sequences from different methods. Each evaluator rated 8 randomly ordered but counter-balanced video pairs comparing our method against baselines on a scale from -2 (strongly prefer baseline) to 2 (strongly prefer ours).

\begin{table*}[t]
\centering
\caption{Quantitative comparison on CelebV-Text dataset. Avg. and Best respectively represent the average and best error among 30 generations.}
\label{tab:main_results}
\renewcommand\arraystretch{1.2}
\resizebox{0.95\linewidth}{!}{
\begin{tabular}{cccccccc}
\hline
Method & \tabincell{c}{Angular Error $\downarrow$ \\ Avg., deg } & \tabincell{c}{Angular Error $\downarrow$ \\ Best, deg} & \tabincell{c}{Correlation Pitch $\uparrow$ \\ Best, deg$^2$} &  \tabincell{c}{Correlation Yaw $\uparrow$ \\ Best, deg$^2$} & \tabincell{c}{AVE $\downarrow$ \\ Avg, deg$^2$}  & \tabincell{c}{Smoothness $\downarrow$ \\ Avg, deg/frame$^3$} & \tabincell{c}{APD $\uparrow$ \\ deg$^2$}  \\
\hline
Constant Head Motion & 26.256 & 26.256 & - & - & 103.283 & - & - \\
Mirror Gaze Inputs & 24.227 & 24.227 & 0.157 & 0.300 & 127.902 & 25.287 & - \\
Ours & \textbf{16.548} & \textbf{10.835} & \textbf{0.509} & \textbf{0.598} & \textbf{89.934} & \textbf{6.987} & \textbf{264.463} \\
\hline
w/o Temporal Modeling & 18.218 & 15.600 & 0.338 & 0.433 & 139.245 & 66.267 & \textbf{435.081} \\
Ours & \textbf{16.548} & \textbf{10.835} & \textbf{0.509} & \textbf{0.598} & \textbf{89.934} & \textbf{6.987} & 264.463 \\
\hline
\end{tabular}}
\vspace{-2mm}
\end{table*}

\begin{table}[t]
\centering
\caption{Human evaluation results. Mean preference scores range from -2 (strongly prefer baseline) to 2 (strongly prefer ours).}
\label{tab:human_eval}
\resizebox{\columnwidth}{!}{
\renewcommand\arraystretch{1.2}
\begin{tabular}{lcccc}
\hline
Baseline to compare with & Mean & Standard deviation & p-value & Sig. \\
\hline
Constant Head Motion & 0.250 & 0.829 & 0.688 & \\
Mirror Gaze Inputs & 1.500 & 0.707 & 0.016 & * \\
w/o Temporal Modeling & 1.625 & 0.484 & 0.008 & ** \\
Driven by Real Motion & 1.000 & 0.707 & 0.031 & * \\
\hline
\multicolumn{5}{l}{* $p < 0.05$, ** $p < 0.01$} \\
\end{tabular}}
\vspace{-2mm}
\end{table}

\subsection{Results and Analysis}
We compare our method against two baseline approaches: \textit{Constant Head Motion}, which maintains fixed head pose throughout the sequence, and \textit{Mirror Gaze Inputs}, which simply copies gaze directions as head poses. These represent simple deterministic approaches to gaze-head coordination.

As shown in Table~\ref{tab:main_results}, our method significantly outperforms these baselines across all metrics. Compared to \textit{Mirror Gaze Inputs}, our approach achieves 55.3\% improvement in best angular error and 72.4\% improvement in smoothness. Our method also shows superior correlation with real head movements in both pitch and yaw dimensions, as well as better AVE. This demonstrates the effectiveness of our data-driven approach in capturing the natural statistical patterns of gaze-head coordination that are difficult to model with handcrafted deterministic rules.

Human evaluation results in Table~\ref{tab:human_eval} further validate our approach, showing statistically significant preferences for our method compared to \textit{Mirror Gaze Inputs} ($p=0.016$, Wilcoxon signed-rank test) and generations driven by real motion ($p=0.031$). The balanced results against \textit{Constant Head Motion} suggest that minimal head movement can be appropriate in some contexts, while our method provides more natural motion when movement is expected.

\begin{figure}[t]
    \centering
    \includegraphics[width=\columnwidth]{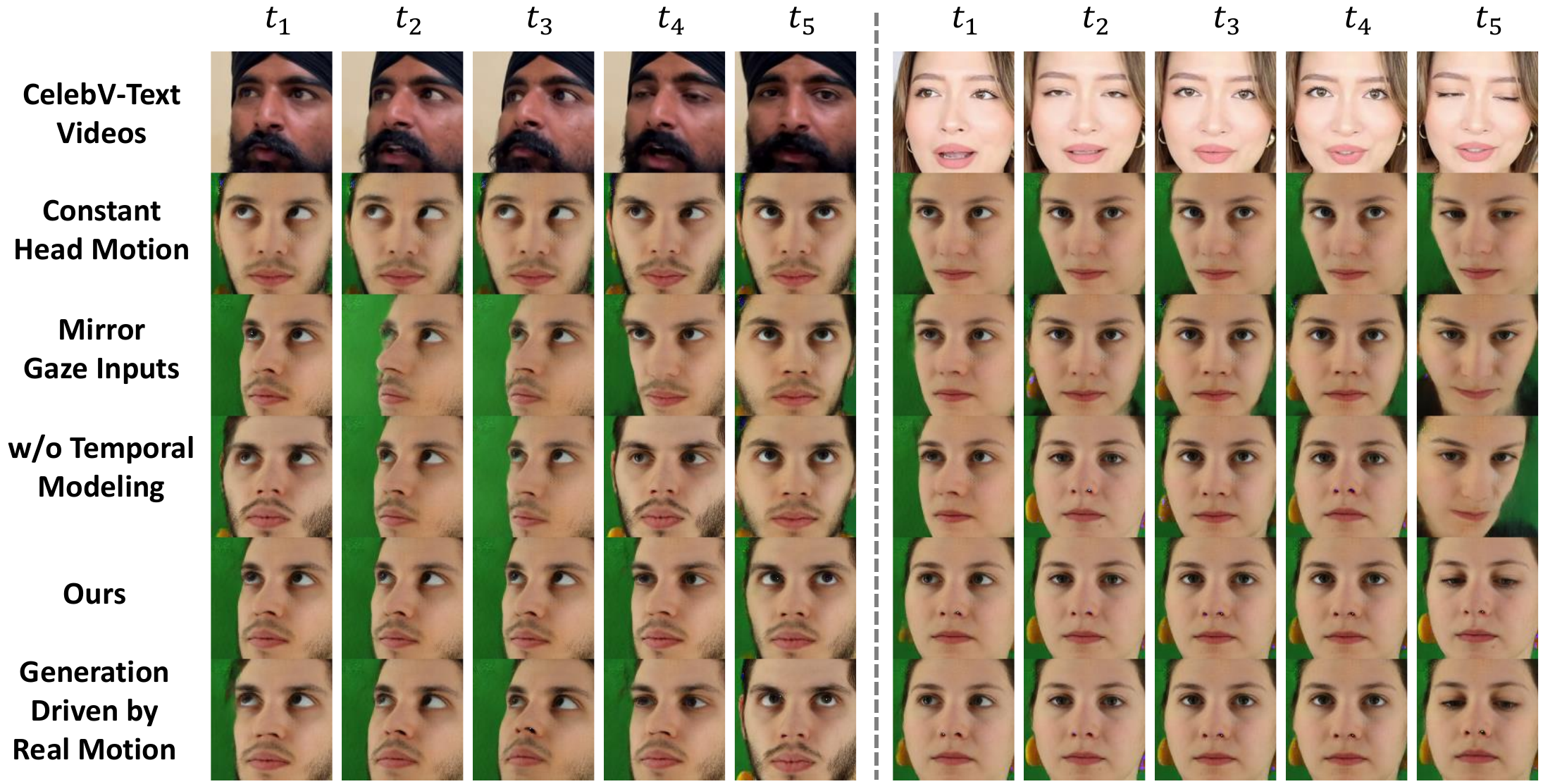}
    \caption{Qualitative comparison of facial videos showing five consecutive frames.}
    \label{fig:quali}
    \vspace{-7mm}
\end{figure}

Qualitative results in Fig.~\ref{fig:quali} visually demonstrate these advantages. Our approach generates natural head movements that appropriately follow gaze direction changes, while \textit{Mirror Gaze Inputs} produces exaggerated head movements and \textit{Constant Head Motion} appears unnatural during gaze shifts.

\subsection{Ablation Study on Temporal Modeling}
To evaluate our temporal modeling approach, we conducted an ablation study by comparing our full model against a variant without temporal components (removing GRU and context conditioning). This comparison isolates the contribution of our temporal modeling approach to the overall performance.

As shown in Table~\ref{tab:main_results}, removing temporal modeling results in significantly degraded performance, with 44.0\% higher best angular error and $8.5 \times$ higher jerk. While the model without temporal constraints achieves higher diversity due to frame-by-frame independence, it does so at the expense of plausibility and natural motion. The correlation scores also drop substantially for both pitch and yaw dimensions, and AVE increases by 54.8\%.

Without temporal modeling, the generated head motions exhibit noticeable jittering and discontinuities between frames, while our full model produces smooth and coherent movements that closely resemble natural human head motion patterns, as shown in Fig.~\ref{fig:quali}.

Human evaluators strongly preferred our full model over this variant, with the highest preference score ($1.625$) and strongest statistical significance ($p=0.008$) among all comparisons. This confirms the critical importance of capturing temporal dynamics for natural gaze-head coordination.

\section{Conclusion}

We present the first data-driven approach for modeling gaze-head coordination, addressing the fundamental challenge of generating natural head movements that complement gaze direction. Our solution captures both the probabilistic nature and temporal dynamics of this coordination through a cVAE framework trained on in-the-wild facial videos. The automated pipeline we developed efficiently extracts diverse training data without specialized hardware, while our model effectively learns the complex distributional and temporal relationships between gaze and head movements. Both quantitative metrics and human evaluation confirm our method's effectiveness, with human evaluators showing strong preference over baseline approaches in generating natural head motions.

While our current approach successfully demonstrates head motion generation coordinated with gaze, integration with facial video generation remains preliminary. Recent advances in facial synthesis~\cite{guo2024liveportrait} suggest opportunities for more realistic outputs that better capture facial expressions alongside head motion. Our future work will focus on end-to-end training of head motion generation with advanced facial synthesis methods to create a unified framework for gaze-controlled facial animation with natural head movements.

\section*{Acknowledgments}
This work was supported by JSPS KAKENHI Grant Number JP22KJ0923. We thank Jiawei Qin for providing the gaze estimation model.

\end{document}